\documentclass[letterpaper, 10 pt, conference]{ieeeconf}  % Comment this line out if you need a4paper

\IEEEoverridecommandlockouts                              % This command is only needed if 
\usepackage[colorlinks,linkcolor=blue]{hyperref}%用于插入超链接
\usepackage{graphicx}
\usepackage{amsmath}
\usepackage{booktabs}
\usepackage{xcolor}
\usepackage{colortbl}
\usepackage{subcaption}
\usepackage{amssymb}

\title{\LARGE \bf
Explore the LiDAR-Camera Dynamic Adjustment Fusion \\for 3D Object Detection
}

\author{Yiran Yang$^{1,*}$, Xu Gao$^{2,*}$, Tong Wang$^{2}$, Xin Hao$^{2}$, Yifeng Shi$^{2}$, Xiao Tan$^{2}$, Xiaoqing Ye$^{2}$, Jingdong Wang$^{2}$% <-this % stops a space
\thanks{This work is supported by Baidu Inc.}% <-this % stops a space
\thanks{$^{1}$Yiran Yang is with the University of Chinese Academy of Sciences and the School of Electronic, Electrical and Communication Engineering, University of Chinese Academy of Sciences, Beijing, China.}
\thanks{$^{2}$Xu Gao, Tong Wang, Xin Hao, Yifeng Shi, Xiao Tan, Xiaoqing Ye, Jingdong Wang are with Baidu Inc, Beijing, China.}
}

\begin{document}

\maketitle
\thispagestyle{empty}
\pagestyle{empty}

\renewcommand{\thefootnote}{\fnsymbol{footnote}}
\footnotetext[1]{Equal contribution.}

%%%%%%%%%%%%%%%%%%%%%%%%%%%%%%%%%%%%%%%%%%%%%%%%%%%%%%%%%%%%%%%%%%%%%%%%%%%%%%%%
\begin{abstract}
Camera and LiDAR serve as informative sensors for accurate and robust autonomous driving systems. However, these sensors often exhibit heterogeneous natures, resulting in distributional modality gaps that present significant challenges for fusion. To address this, a robust fusion technique is crucial, particularly for enhancing 3D object detection.  In this paper, we introduce a dynamic adjustment technology aimed at aligning modal distributions and learning effective modality representations to enhance the fusion process. Specifically, we propose a triphase domain aligning module. This module adjusts the feature distributions from both the camera and LiDAR, bringing them closer to the ground truth domain and minimizing differences. Additionally, we explore improved representation acquisition methods for dynamic fusion, which includes modal interaction and specialty enhancement. Finally, an adaptive learning technique that merges the semantics and geometry information for dynamical instance optimization. Extensive experiments in the nuScenes dataset present competitive performance with state-of-the-art approaches. Our \href{https://github.com/youngfly/Dynamic-Adjustment-Fusion}{code} will be released in the future.

\end{abstract}

%%%%%%%%%%%%%%%%%%%%%%%%%%%%%%%%%%%%%%%%%%%%%%%%%%%%%%%%%%%%%%%%%%%%%%%%%%%%%%%%
\section{INTRODUCTION}
\label{sec:intro}

With the advancement of autonomous driving, the 3D object detection task has gained significant attention as a crucial component of environmental perception. Consequently, vehicles are typically equipped with various sensors, including multi-view cameras and LiDAR. These two sensor types provide abundant and diverse input information, encompassing RGB data and point clouds. Specifically, RGB images offer rich semantic information, while point clouds provide geometric constraints. Vision-based strategies \cite{wang2021detr3d, li2022bevdepth, huang2022bevdet4d, liu2022petr, liu2022petrv2, li2022bevformer, yang2022bevformer} excel at classifying different objects but may suffer from inaccurate localization. In contrast, LiDAR-based approaches, exemplified by works like \cite{yan2018second, lang2019pointpillars, wang2020pillar} effectively locate objects but may exhibit classification inaccuracies. The central challenge lies in fusing these two modalities as complementary sources to achieve precise and robust object detection.

Camera and LiDAR sensors typically exhibit distinct feature distributions, and early fusion approaches follow a ‘from-camera to LiDAR’ strategy, which can be categorized into three schools. Some methods adopt point-level fusion strategies. For instance, PointPainting \cite{vora2020pointpainting}  and PointAugmenting \cite{wang2021pointaugmenting} overlay image information onto each LiDAR point cloud, enhancing feature representation. Other approaches focus on feature-level fusion, such as DeepFusion \cite{li2022deepfusion} , AutoAlignV2  \cite{chen2022autoalignv2}, and Graph R-CNN \cite{yang2022graph}. Additionally, some methods fuse information at the proposal level, including TransFusion \cite{bai2022transfusion} and FUTR3D \cite{chen2022futr3d}. However, these LiDAR-dominant fusion strategies face two challenges: (1) Camera and LiDAR features exhibit significantly different densities, resulting in fewer camera features being matched to a LiDAR point (especially for 32-channel LiDAR scanners). (2) These methods are sensitive to sensor misalignment due to the rigid association between points and pixels established by calibration matrices.

\begin{figure}[t]
    \centering
    \includegraphics[scale=0.27]{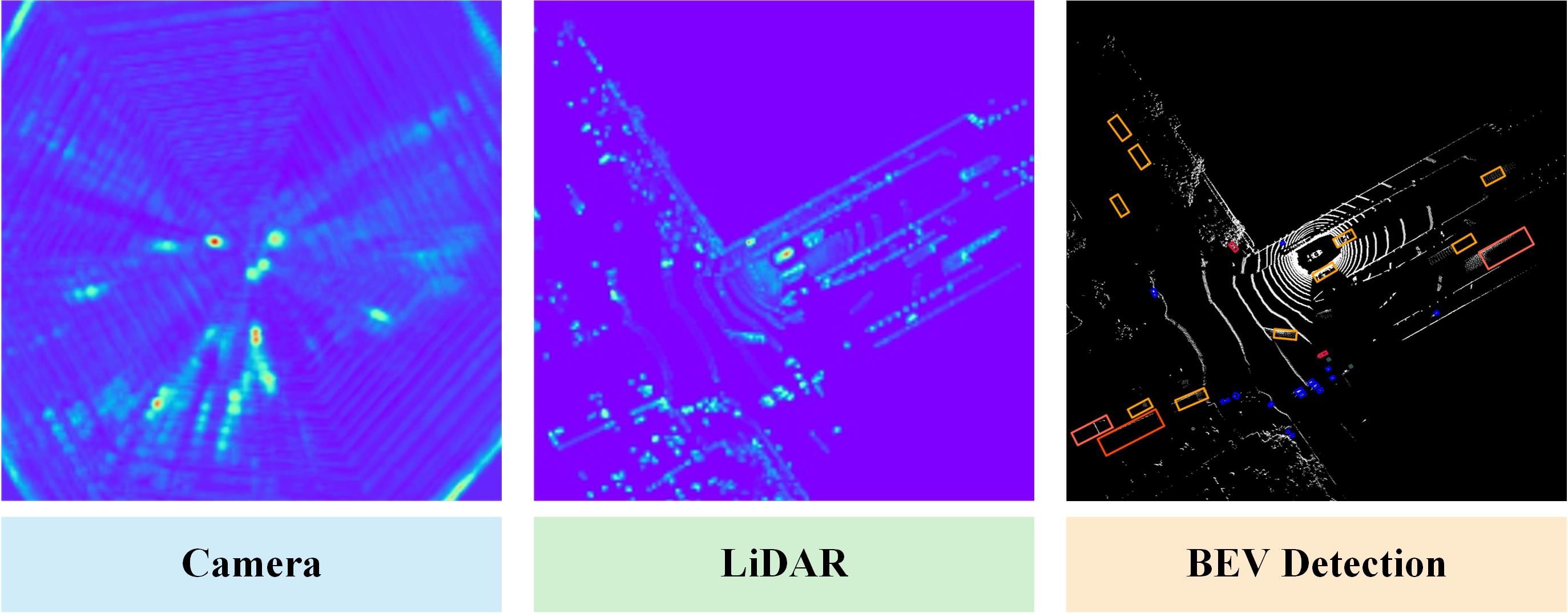}
    \caption{The visualization of BEV features and detection results. Diverse modality usually have diverse perception ability.}
    \label{feature}
\vspace{-10pt}
\end{figure}

\begin{figure*}
\centering
\includegraphics[scale=0.40]{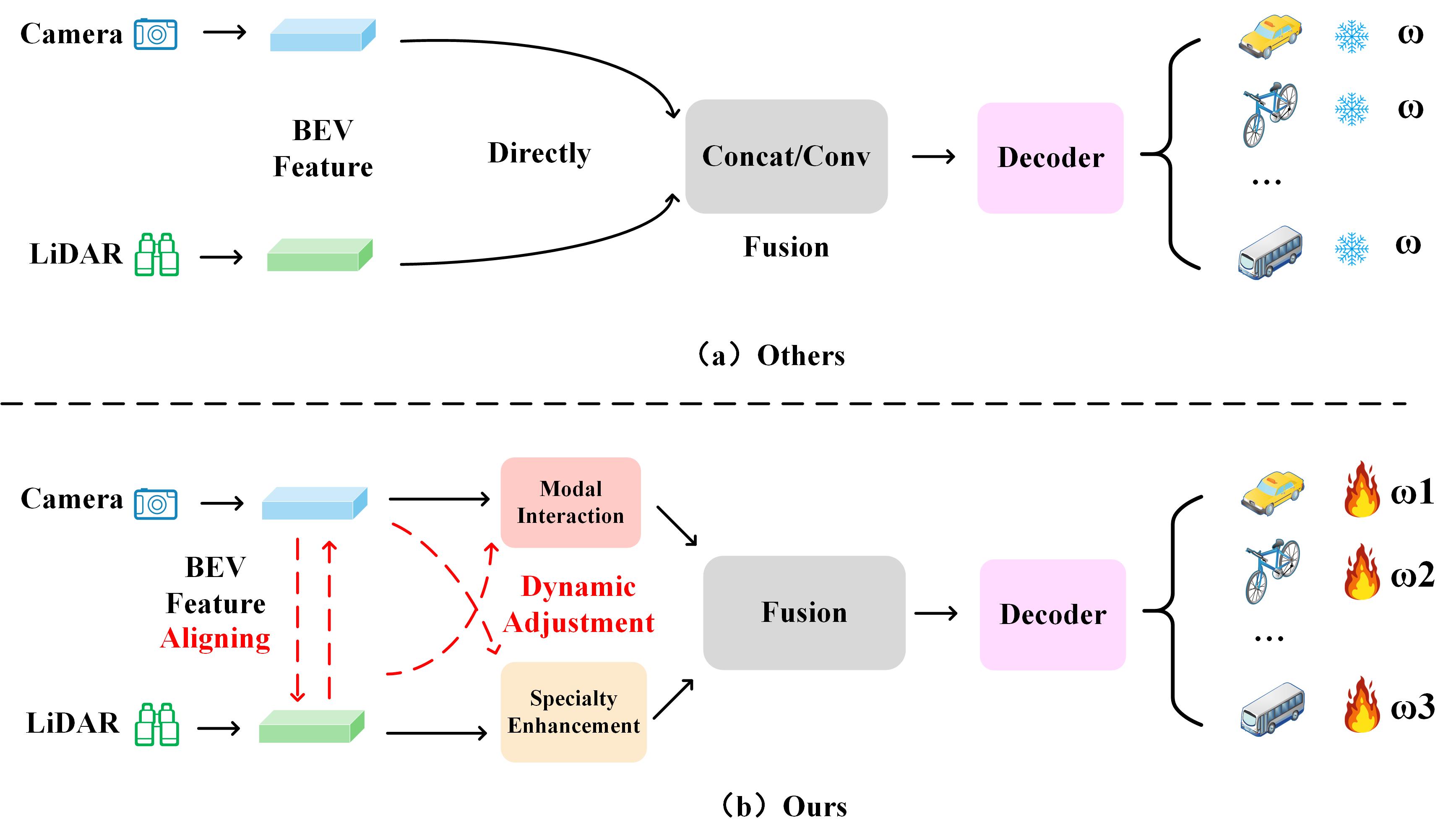}
\caption{Comparison between existing multi-modal fusion and our strategy. (a) In other methods, each subnet encodes each modal feature and then fuses directly. (b) We propose to adopt modal aligning and dynamical adjustment to get better representations and fuse them adaptively by channel and space. Moreover, we use a dynamic technique to optimize instance.}
\label{issue}
\end{figure*}

In recent years, the joint fusion of camera and LiDAR features (depicted in Fig. 1(a)) has replaced early single-modal dominant strategies. Some approaches adopt bidirectional interactions between the two modalities to achieve deep fusion, as demonstrated by DeepInteraction \cite{yang2022deepinteraction}. Meanwhile, others \cite{liang2022bevfusion, liu2022bevfusion} construct a unified bird’s-eye-view representation space to fuse different modal features. Overall, most 3D multi-modal object detection methods focus on developing sophisticated fusion mechanisms that span from single-modal dominance to multi-modal joint fusion. Despite these impressive advancements, these strategies are often affected by heterogeneous modality gaps. As illustrated in Fig. \ref{feature}, multi-modal sensors yield diverse feature patterns, each with varying perception abilities toward the environment. Therefore, learning latent modality representations and capturing crucial modal properties are efficient ways to facilitate multi-modal fusion. To achieve this goal, we explore dynamic adjustment fusion between LiDAR and camera data, which effectively enhances each modality’s representation and fuses complementary information for improved 3D object detection.

Inspired by recent advancements in multi-modal fusion approaches, we propose to explore the dynamic adjustment fusion technique (depicted in Fig. \ref{issue} (b)). This technique learns subspaces for each modality and explores the relevance between two modalities, resulting in improved representations for fusion. Before delving into representation learning, we design a triphase domain alignment module that aligns two modalities with each other, bringing their space distributions closer to the ground truth domain. To enhance modality representation and capture key properties, we devise a modal interaction module that explores the relevance between camera and LiDAR modalities, improving correlated representation. Additionally, we investigate their specific perception of object regions to enhance each modality’s specialty. Finally, we adopt a dynamic fusion strategy that combines the aforementioned interaction and specialty representations across spatial and channel dimensions. Furthermore, recognizing that different objects exhibit diverse visual sizes, we propose an adaptive learning technique that dynamically optimizes instances based on semantics and geometry, rather than treating them equally. Experiments conducted on the nuScenes dataset \cite{caesar2020nuscenes} demonstrate competitive performance compared to state-of-the-art methods. Our contributions are summarized as follows:

\begin{itemize}
	\item[$\bullet$]
	We propose a novel framework to explore the LiDAR-camera dynamic adjustment fusion. Massive experiments on the nuScenes benchmark prove our effectiveness. 
		
	\item[$\bullet$]
	For the multi-modal fusion, we first design a triphase domain aligning module to learn domain-adaptive feature representations. Second, the modal interaction and specialty enhancement module dynamically improve the representation. Finally, the dynamic fusion process yields a high-quality fusion representation based on the aforementioned steps.
			
	\item[$\bullet$]
    For instance optimization, we propose an adaptive learning technique that dynamically optimizes diverse instances by combining semantic and geometric information.
\end{itemize}

\begin{figure*}
    \centering
    \includegraphics[scale=0.09]{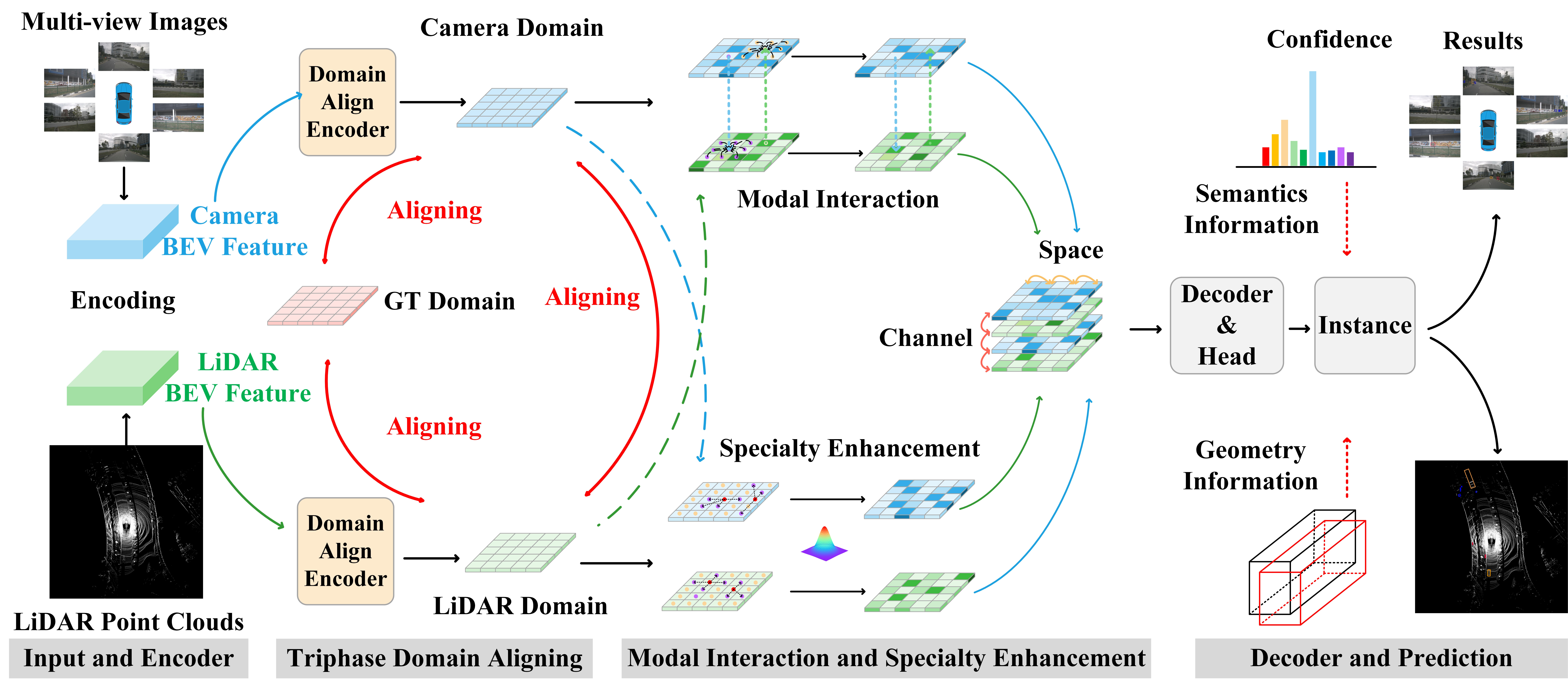}
    \caption{The illustration of our framework. Firstly, multi-modal features are extracted by each encoder and aligned by a triphase domain aligning module to adjust feature distributions. Then, we explore the modal interaction and specialty enhancement to get better representations for dynamic fusion. An adaptive learning technique fuses semantics and geometry information to adaptively optimize instances. The model decodes fused features and predicts results finally.}
	\label{framework}
\end{figure*}

\section{Related Work}
\subsection{Single-modal 3D object detection}
Automated driving vehicles are typically equipped with multiple sensors. However, in the early stages of 3D object detection, most approaches rely on single-modal data from either cameras or LiDAR. Camera-based methods can be broadly categorized into two schools: monocular detection and multi-view detection. The KITTI benchmark \cite{geiger2012we} primarily features a single front camera, and many methods \cite{park2021pseudo, zhou2021monocular, liu2021autoshape,liu2022learning, wang2021fcos3d, wang2021probabilistic} initially focused on monocular detection. Nevertheless, with the emergence of large-scale autonomous driving datasets like nuScenes \cite{caesar2020nuscenes} and Waymo \cite{sun2020scalability}, multi-view input data have become increasingly important, providing richer information and driving a new trend in the field. Inspired by DETR \cite{carion2020end} and Lift-Splat-Shoot \cite{philion2020lift}, an increasing number of multi-view detectors have emerged. DETR3D \cite{wang2021detr3d} is the first to introduce transformers for end-to-end 3D detection. PETR \cite{liu2022petr, liu2022petrv2} leverages position embeddings to create 3D position-aware features, enhancing object localization. BEVDet \cite{huang2021bevdet, huang2022bevdet4d}, BEVDepth \cite{li2022bevdepth}, and BEVFormer \cite{li2022bevformer, yang2022bevformer} transform 2D features into bird’s-eye view (BEV) representations, enabling object detection in a unified BEV space. PETRV2 \cite{liu2022petrv2}, BEVDet4D \cite{huang2022bevdet4d}, and BEVFormer \cite{li2022bevformer} also incorporate temporal cues for impressive performance gains. Additionally, LiDAR-based approaches can be categorized into three classes: point-based methods \cite{li2021LiDAR, qi2018frustum, qi2017pointnet, qi2017pointnet++, shi2019pointrcnn, yang20203dssd}, which directly process raw LiDAR point clouds; voxel-based methods \cite{zhou2018voxelnet}, which transform points into a 3D voxel grid; and pillar-based methods \cite{lang2019pointpillars, wang2020pillar}, which extract features similar to CNNs from point pillars
	
\subsection{Multi-modal 3D object detection}
Multi-modal fusion can assemble advantages in each data from different sensors. Recently, existing fusion methods have been classified into two following approaches. Camera-to-LiDAR methods usually project the camera features to LiDAR features and finish the fusion, which means the LiDAR is dominant. PointPainting \cite{vora2020pointpainting} paints segmentation scores onto each point in the LiDAR point cloud. PointAugmenting \cite{wang2021pointaugmenting} paints features from the 2D image onto each point in the LiDAR point cloud. DeepFusion \cite{li2022deepfusion} proposes Inverse-Aug and Learnable-Align to better fuse camera and LiDAR modalities. AutoAlignV2 \cite{chen2022autoalignv2} employs a deformable feature aggregation module, which attends to sparse learnable sampling points for cross-modal relational modeling. Graph R-CNN \cite{yang2022graph} utilizes a dynamic point aggregation strategy, which sampled context and object points, and visual features augmentation to decorate the points with 2D features. Transfusion \cite{bai2022transfusion} employs a soft-association mechanism to finish the LiDAR and camera fusion which handles inferior image conditions. FUTR3D \cite{chen2022futr3d} introduces the first unified end-to-end sensor fusion framework, which can be used in almost any sensor configuration. MSMDFusion \cite{jiao2022msmdfusion} encourages sufficient LiDAR-camera feature fusion in the multiscale voxel space. 

Recently, joint fusion between cameras and LiDAR has demonstrated significant effectiveness. Two BEVFusion approaches \cite{liang2022bevfusion, liu2022bevfusion} project camera and LiDAR features into bird’s-eye view space, enabling object detection with a unified fused representation. Additionally, DeepInteraction \cite{yang2022deepinteraction} introduces a strategy where individual per-modality representations are learned and maintained, preserving their unique characteristics. 

\begin{figure*}
    \centering
    \includegraphics[scale=0.15]{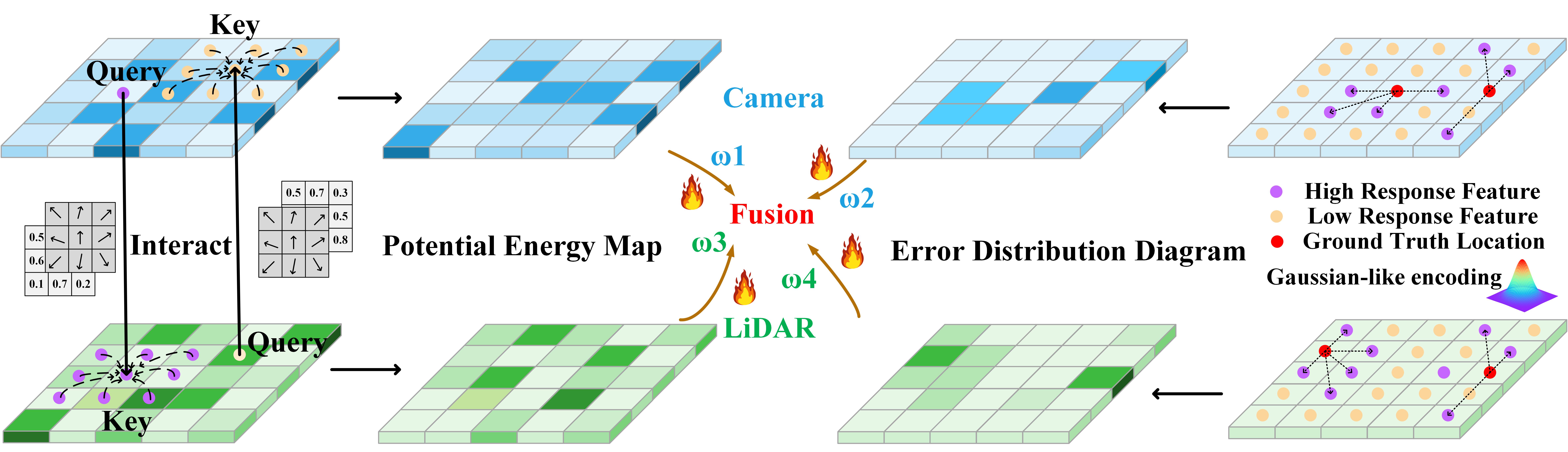}
    \caption{The illustration of modal interaction and specialty enhancement. The left part is the modal interaction and the right part is the modal specialty enhancement. We fuse the representations dynamically.}
    \label{MML}
\end{figure*}

\section{Dynamic Adjustment Fusion}
\subsection{Overview}
Drawing upon the high inference speed and transformation flexibility, we adopt BEVFusion \cite{liu2022bevfusion}, a state-of-the-art 3D object detection method, as our baseline. BEVFusion leverages bird’s-eye-view (BEV) representation for multimodal fusion. However, the basic fusion strategy in the baseline, which relies on convolution operations, is overly simplistic for effectively fusing complex features. To address this limitation, we propose a novel fusion framework, illustrated in Fig. \ref{framework}. Let’s delve into its four key components: First, we encode RGB information from multi-view cameras and point clouds from LiDAR, resulting in BEV features for both modalities. To address feature mismatch between diverse modalities, we design a triphase aligning module that adjusts feature distributions and aligns them in both spatial and channel dimensions. Next, our modal interaction and specialty enhancement module effectively fuses complementary information while reducing redundancy. We then apply dynamic fusion to adjust features in both spatial and channel dimensions. Additionally, we employ an adaptive learning technique to optimize diverse instances during training. Finally, predictions are completed.

% Based on the high inference speed and transformation flexibility, BEVFusion \cite{liu2022bevfusion}, a state-of-the-art 3D object detection method, is adopted as our baseline. BEVFusion utilizes the bird's-eye-view (BEV) as the unified representation for multimodal fusion. However, the basic fusion strategy in baseline, which uses a convolution operation, is too simple to better fuse the complex features. Aiming to solve the above fusion issue, we propose a novel fusion framework, which is illustrated in Fig. \ref{framework}. Four parts are introduced in detail below. First, We encode RGB information from multi-view cameras and point clouds from LiDAR, then get BEV features in two modalities. To avoid the feature mismatch between diverse modalities. We first design a triphase aligning module, which adjusts the feature distributions and align them in space and channel level. Next, we design a modal interaction and specialty enhancement module to validly fuse the complementary information to reduce redundancy. Then, a dynamic fusion is applied to fuse adjustment features in the space and channel dimensions. Moreover, an adaptive learning technique is used to optimize diverse instances in the training process. Finally, predictions is completed. 

\subsection{Triphase Domain Aligning}
As depicted in Fig. \ref{feature}, different sensor modalities yield distinct feature patterns. On one hand, they observe different regions of objects. On the other hand, their feature distributions vary due to different encoding patterns. Although previous fusion strategies focus on feature-level aggregation, domain mismatch issues still persist in feature fusion.
	
Before fusing multimodal features, we align them to a common domain. Specifically, we define the camera domain as $\Psi(C)$, the LiDAR domain as $\Psi(L)$, and the ground truth domain as $\Psi(G)$. Directly optimizing them to a shared domain can be challenging, so we introduce feature-level constraints. We employ domain aligning encoders to process the original bird's-eye-view (BEV) features, resulting in camera BEV aligning features denoted as $F_x$ and LiDAR BEV aligning features denoted as $F_y$. These aligning encoders can be implemented using convolution layers or transformer structures.
 
To align $\Psi(C)$ and $\Psi(L)$, we employ a basic $\mathcal{L}_1$ constraint on $F_x$ and $F_y$. However, solely matching $\Psi(C)$ and $\Psi(L)$ is insufficient, as it may deviate from the ground truth domain $\Psi(G)$. To address this, we introduce a $GaussianFocal$ constraint to ensure that $\Psi(C)$ and $\Psi(L)$ are appropriately aligned. The overall triphase domain aligning optimization is formulated as follows:
\begin{equation}
\begin{aligned}
\mathcal{L}_{t} =  \lambda_{1}\mathcal{L}_1(F_x, F_y) + \lambda_{2}\sum\limits_{z \in \{ x,y\} } {\mathcal{L}_{GFL}({F_z},{F_g})},
\end{aligned}
\end{equation}
Where $\lambda_{1}$ and $\lambda_{2}$ represent coefficients for balancing. The heatmap supervision $F_g$ is generated from ground truth, similar to the approach used in CenterNet [1]. We create a bird's-eye-view (BEV) space and apply a Gaussian kernel to each center point to obtain the heatmap supervision $F_g$. The loss function $\mathcal{L}_{GFL}$ is based on Gaussian focal loss, also inspired by CenterNet \cite{zhou2019objects}. It combines a Gaussian kernel with focal loss to supervise the feature map.
	
While the constraint between two modalities is designed to align their feature distributions, our proposed triphase domain aligning strategy ensures a balance between feature alignment and feature complementarity. The heatmap supervision is tailored for both cameras and LiDAR features to maintain their complementary representations. Additionally, we set the weight ratio between the constraint of the two modalities and the heatmap supervision as 1:10. This ensures that the constraint on the modalities does not compromise their unique expressions, preserving complementarity.

\subsection{Modal Interaction and Specialty Enhancement}
Camera and LiDAR information are captured from different sensors and encoded by specific encoders. Existing approaches often fuse different features using simple “concat/conv” operations, but the unique advantages of each modality are not fully exploited. To address this, we propose a modal interaction and specialty enhancement approach that leverages the full potential of each modality. The entire process is illustrated in Fig. \ref{MML}.

\subsubsection{Modal Interaction}
In this section, we begin by performing modal interaction to capture relevance and enhance similar representations. To obtain the relationship between the camera and LiDAR modalities, we utilize the deformable transformer \cite{zhu2020deformable}, denoted as $\Phi$. The formulation is as follows:
\begin{equation}
    \begin{aligned}
        \Phi(q,p^*,r)= \sum\limits_{m = 1}^M {{W_m}\sum\limits_{k = 1}^K{{A_{mk}}W_m^*} } r(p^* + \Delta {p^*_{mk}}),
    \end{aligned}
\end{equation}
where $q$, $p^*$, and $r$ denote the query, reference point and input feature, respectively. $m$ represents the attention head and $k$ indexs the sampled key. $W$ and $W^*$ are learnable weights. $\Delta {p^*_{mk}}$ and $A_{mk}$ denote the sampling offset and attention weight of the $k^{th}$ sampling point in the $m^{th}$ attention head, respectively. 

Unlike previous direct interactions between features, we utilize heatmaps generated by features to create potential energy maps. These potential energy maps then act on modal features to enhance similar representations. Our strategy avoids disrupting feature distributions, leading to improved learning outcomes. First, we obtain the normalized heatmaps $F_x^\ast$ and $F_y^\ast$. Given camera queries $F_x^\ast$ and LiDAR queries $F_y^\ast$, we model the relevance between them using similar feature encoding, as follows:
\begin{equation}
    \begin{aligned}
        P_{x}^{\ast}=  \Phi(F_x^\ast, p^\ast, F_y^\ast),
        P_{y}^{\ast} = \Phi(F_y^\ast, p^\ast, F_x^\ast).
    \end{aligned}
\end{equation}
Additionally, we model self-interaction using $H_{x}^{\ast}$ and $H_{y}^{\ast}$, where each modality serves as both queries and values. Consequently, the potential energy map for modal interaction representation is defined as $ P_{z} = \text{Conv}(\text{Concat}(P_{z}^{\ast}, H_{z}^{\ast})), \quad z \in \{x, y\}$.

\subsubsection{Specialty Enhancement}
In addition to exploring modal interaction, we delve into their specific representations to enhance their modal specialties. As depicted in Fig. \ref{feature}, cameras and LiDAR typically observe different objects in their feature maps, implying distinct preferences. Therefore, understanding their modal specialties complements the focus on similar latent features. Furthermore, we observe that object regions exhibit higher responses than the background in the feature map. Additionally, the response amplitude gradually decreases with distance from the object. This suggests that regions with high responses have low center offsets from the ground truth and low uncertainty. Inspired by this observation, we model features as Gaussian-like distributions, where points with low offset and low uncertainty exhibit heightened perceptual significance. Specifically, we encode the bird's-eye-view (BEV) feature map into two representations: one denoting the offset $\mu$ from each feature point to the nearest object, and another denoting the uncertainty estimate $\varepsilon$ for that point. We employ basic convolutional neural networks (CNNs) to complete the encoding for both camera and LiDAR modalities. As illustrated in Fig. \ref{MML}, we obtain error distribution diagrams $E$ for each modality.
\begin{equation}
    \begin{aligned}
        E_z = \frac{{\rm{1}}}{\varepsilon_z }\exp ( - \frac{{{\mu_z ^2}}}{{{\varepsilon_z ^2}}}), z \in \{ x,y\}. 
    \end{aligned}
\end{equation}
The offset $\mu$ and uncertainty $\varepsilon_z$ optimization is presented below,
\begin{equation}
    \begin{aligned}
        \mathcal{L}_{s} = {\raise0.7ex\hbox{$\zeta$} \!\mathord{\left/
                {\vphantom {\zeta  N}}\right.\kern-\nulldelimiterspace}
            \!\lower0.7ex\hbox{$N$}}\left( {\sum\limits_{z  \in \{ {x},{p}\} } {\left\| {\mu_z  - {\mu ^*}} \right\|_2^2 +  {\left\| \varepsilon_z  \right\|_2^2} } } \right),
    \end{aligned}
\end{equation}
where $\mu^*$ is the target offset from each feature point to its nearest ground truth. $\zeta$ is the coefficient to balance loss scale. The regularization for uncertainty $\varepsilon_z$  can avoid trivial solutions and encourage the model to be optimistic about accurate predictions. $N$ is the number of feature points. The $E_z$ represent different modal perception ability and can adjust the diverse modal importance in following the fusion process.

\subsubsection{Dynamic Fusion}
\label{DF}
After generating the modal interaction and specialty representations $P$ and $E$, we fuse this information with each modality. We employ a selective kernel network to adjust channel attention. Finally, we fuse the camera and LiDAR modalities using a convolutional network and send the fusion feature to the decoder and head for downstream predictions.

\subsection{Adaptive Learning Technique}
In the context of automated driving scenarios, different objects typically exhibit varying visual sizes. To address the challenge of static optimization for diverse objects, we propose an adaptive learning technique for instance optimization. The prediction of an instance typically includes its category and other 3D attributes. A high-quality instance is characterized by both high confidence and accurate location. To quantify instance quality, we combine the instance's classification score, denoted as \(c(q)\), with the predicted Intersection over Union (\(IoU(q)\)) as an index: \(\varphi(q) = c(q) \times IoU(q)^\eta\). Here, \(\eta\) represents the coefficient for balancing each index. The higher the \(\varphi(q)\), the better the quality of instance \(q\). In practice, we use \(\varphi(q)^* = e^{\varphi(q)}\) as the learning weight for instance optimization.

Following previous work, we adopt Focal loss \cite{lin2017focal}, $L1$ loss, and Gaussian Focal loss \cite{zhou2019objects} for classification, 3D bounding box location, and heatmap supervision, respectively. And the optimization is shown below, 
\begin{equation}
	\begin{aligned}
		\mathcal{L} =& \sum\limits_{q \in {Q^{\rm{*}}}} {\alpha \varphi (q)^*{\mathcal{L}_{cls}}({c_q},c_q^*)} {\rm{ + }}\beta \varphi (q)^*{\mathcal{L}_{loc}}({b_q},b_q^*)\\
		&+\gamma\mathcal{L}_{heat}(\mathcal{H}, \mathcal{H}^*) + \mathcal{L}_{s} + \mathcal{L}_{tda},
	\end{aligned} 
	\label{opt_new}
\end{equation}
where $\alpha, \beta, \gamma$ are coefficients to balance the loss. $Q^*$ are the queries (instances) matched to the ground truth. $c_q, b_q,\mathcal{H}$ are the predictions of classification, location, and heatmap, respectively. And $c_q^*, b_q^*, \mathcal{H}^*$ are corresponding ground truth. $\mathcal{L}_{s}, \mathcal{L}_{t}$ are specialty enhancement module and triphase domain aligning optimization, respectively. 

\begin{table*}[t]
    \centering
    \caption{NuScenes $test$ set evaluation with SOTA approaches.}
    \resizebox{17cm}{!}{
        \setlength{\tabcolsep}{1.7mm}
        {
            \begin{tabular}{c|c|cc|cccccccccc}
                \toprule
                Method & Modality& mAP &NDS &Car &Truck &C.V. &Bus &Trailer &Barrier &Motor. &Bike &Ped. &T.C.\\
                \midrule
                PointPillar \cite{lang2019pointpillars} &L &40.1 &55.0 &76.0 &31.0 &11.3 &32.1 &36.6 &56.4 &34.2 &14.0 &64.0 &45.6 \\
                CenterPoint \cite{yin2021center} &L &60.3 &67.3 &85.2 &53.5 &20.0 &63.6 &56.0 &71.1 &59.5 &30.7 &84.6 &78.4\\
                TransFusion-L \cite{bai2022transfusion}  &L &65.5 &70.2 &86.2 &56.7 &28.2 &66.3 &58.8 &78.2 &68.3 &44.2 &86.1 &82.0\\
                \midrule
                PointPainting \cite{vora2020pointpainting}  &LC &46.4 &58.1 &77.9 &35.8 &15.8 &36.2 &37.3 &60.2 &41.5 &24.1 &73.3 &62.4 \\
                MVP \cite{yin2021multimodal} &LC &66.4 &70.5 &86.8 &58.5 &26.1 &67.4 &57.3 &74.8 &70.0 &49.3 &89.1 &85.0\\
                PointAugmenting \cite{wang2021pointaugmenting} &LC &66.8 &71.0 &87.5 &57.3 &28.0 &65.2 &60.7 &72.6 &74.3 &50.9 &87.9 &83.6 \\
                UVTR \cite{li2022unifying}  &LC &67.1 &71.1 &87.5 &56.0 &33.8 &67.5 &59.5 &73.0 &73.4 &54.8 &86.3 &79.6\\
                VFF \cite{li2022voxel}  &LC &68.4 &72.4 &86.8 &58.1 &32.1 &70.2 &61.0 &73.9 &\textbf{78.5} &52.9 &87.1 &83.8\\
                TransFusion \cite{bai2022transfusion} &LC &68.9 &71.7 &87.1 &60.0 &33.1 &68.3 &60.8 &78.1 &73.6 &52.9 &88.4 &86.7\\
                BEVFusion \cite{liang2022bevfusion}  &LC &69.2 &71.8 &88.1 &60.9 &34.4 &69.3 &62.1 &78.2 &72.2 &52.2 &89.2 &85.2\\
                BEVFusion \cite{liu2022bevfusion}  &LC &70.2 &72.9 &88.6 &60.1 &\textbf{39.3} &69.8 &63.8 &80.0 &74.1 &51.0 &89.2 &86.5\\
                DeepInteraction \cite{yang2022deepinteraction}  &LC &70.8 &73.4 &87.9 &60.2 &37.5 &70.8 &63.8 &\textbf{80.4} &75.4 &54.5 &\textbf{90.3} &87.0\\
                Ours &LC &\textbf{71.8} &73.2 &\textbf{89.2} &\textbf{63.4} &38.5 &\textbf{74.7} &\textbf{66.6} &78.7 &75.4 &\textbf{54.8} &89.9 &86.6 \\
                \bottomrule
    \end{tabular}}}%
    \label{test}%
\end{table*}%

\begin{table}
	\centering
 	\caption{NuScenes $val$ set evaluation with SOTA approaches.}
	\resizebox{7.9cm}{!}{
		\setlength{\tabcolsep}{2.6mm}
		{
			\begin{tabular}{c|c|cc}
				\toprule
				Method & Modality& mAP &NDS \\
				\midrule
				PETR \cite{liu2022petr} &C &37.0 &44.2 \\
				BEVFormer \cite{li2022bevformer} &C &41.6 &51.7 \\
				\midrule
				FUTR3D \cite{chen2022futr3d}  &L &59.3 &65.5 \\
				UVTR \cite{li2022unifying}   &L &60.8 &67.6 \\
				TransFusion-L \cite{bai2022transfusion}  &L &65.5 &70.2 \\
				\midrule
				FUTR3D \cite{chen2022futr3d}  &LC &64.5 &68.3 \\
				UVTR \cite{li2022unifying}   &LC &65.4 &70.2 \\
				PointPainting \cite{vora2020pointpainting} &LC &65.8 &69.6 \\
				FusionPainting \cite{xu2021fusionpainting} &LC &66.5 &70.7 \\
				AutoAlign \cite{chen2022autoalign} &LC &66.6 &71.1 \\
				TransFusion \cite{bai2022transfusion}  &LC &67.5 &71.3 \\
				BEVFusion \cite{liang2022bevfusion}  &LC &67.9 &71.0 \\
				BEVFusion \cite{liu2022bevfusion}  &LC &68.5 &71.4 \\
				DeepInteraction \cite{yang2022deepinteraction} &LC &69.9 &72.6\\
				Ours &LC &\textbf{71.0} &\textbf{72.8} \\
				\bottomrule
	\end{tabular}}}%
	\label{val}%
\end{table}%
 
\section{Experiments}
\subsection{Experimental Setup}
\noindent\textbf{Dataset:} We adopt the nuScenes dataset \cite{caesar2020nuscenes} to evaluate our approach. The benchmark contains multi-modal data that are collected from 6 cameras, 1 LiDAR, and 5 radars. There are 1000 scenes and is officially divided into 700/150/150 scenes as the training/validation/test set, respectively.

\noindent\textbf{Metric:} We also follow the official evaluation metric including the mean average precision (mAP) and nuScenes detection score (NDS). The mAP is calculated by averaging over distance thresholds of 0.5m, 1m, 2m, and 4m across 10 categories. NDS is a consolidated metric that is calculated by mAP, mean Average Translation Error (mATE), mean Average Scale Error (mASE), mean Average Orientation Error (mAOE), mean Average Velocity Error (mAVE), and mean Average Attribute Error (mAAE).

\subsection{Implementation Details}
We use the Swin-Tiny \cite{liu2021swin} and VoxelNet \cite{zhou2018voxelnet} as the image and LiDAR backbone, respectively. Voxel size is (0.075m, 0.075m, 0.2m) and the detection range is set to [-54m, 54m] for $X$ and $Y$ axis and [-5m, 3m] for $Z$ axis. And only single-level fused features are used for prediction. Our model training process includes two stages. 1) We first train the camera model and LiDAR model with each modality data with 20 epochs, respectively. 2) Then we train the fusion model for 6 epochs and inherit weights from two trained streams. Our model is trained with a batch size of 16 on 8 V100 GPUs. AdamW optimizer is adopted for optimization. The CosineAnnealing learning rate policy is used and the initial learning rate is $1 \times 10^{-4}$. We also set $\lambda_{1}=0.1$, $\lambda_{2}=\rho_{1}=\rho_{2}=\eta=1$, $\zeta=2000$, $\alpha=1$, $\beta=0.25$, $\gamma=1$.

\subsection{State-of-the-art Comparison}
\begin{figure}
    \centering
    \includegraphics[scale=0.15]{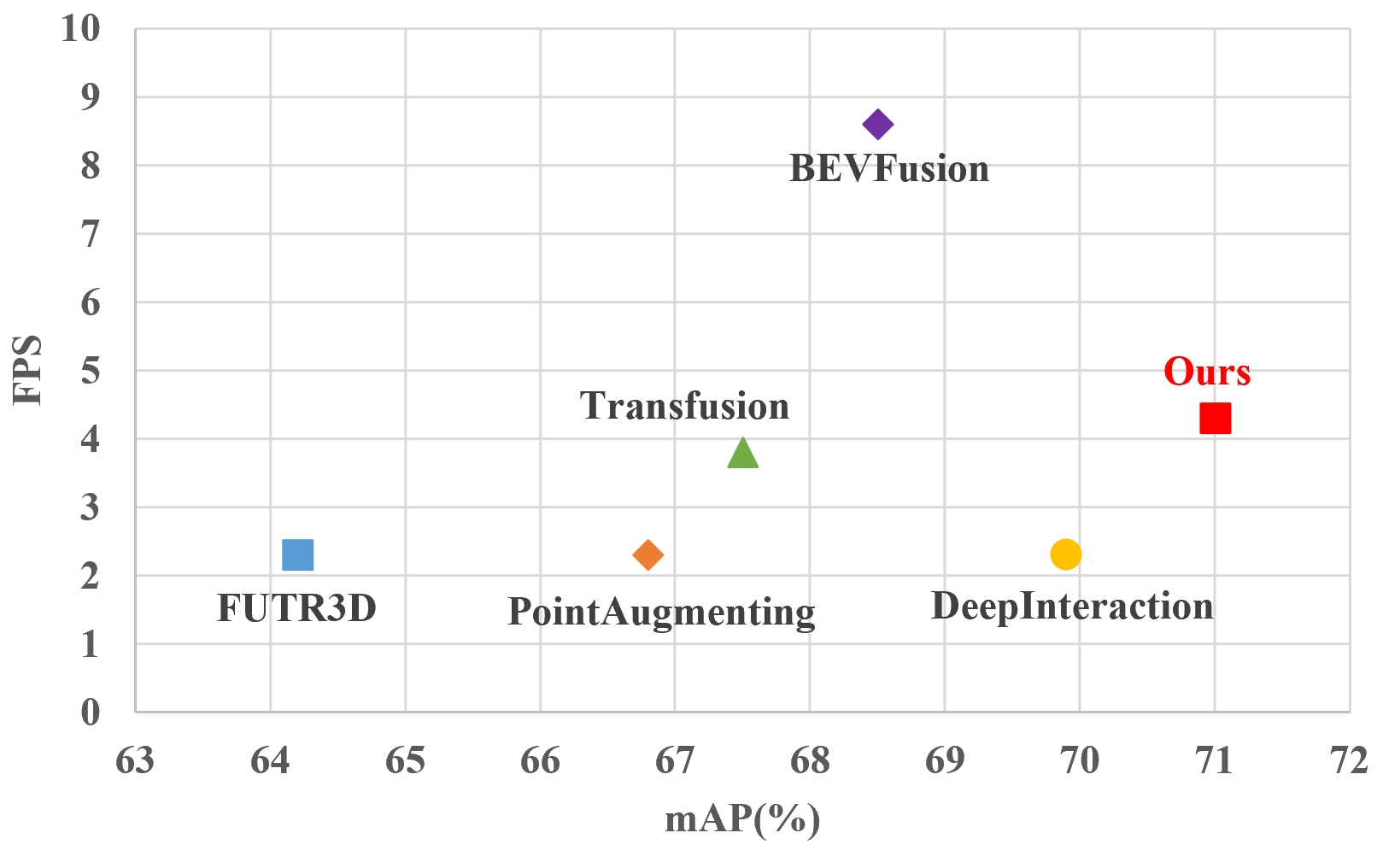}
    \caption{Inference speed comparison.}
    \label{FPS}
\end{figure}

In this section, we present results from the validation and test sets on the nuScenes benchmark. As shown in Table \ref{test}, our method demonstrates competitive performance compared to other fusion strategies and LiDAR-based approaches, achieving 71.8\% mAP and 73.2\% NDS. Additionally, in the validation set (Table \ref{val}), our model achieves impressive results with a 71.0\% mAP and 72.8\% NDS. These outcomes highlight the performance gains resulting from our three novel design proposals. Furthermore, Fig. \ref{FPS} illustrates that our method achieves higher accuracy and lower latency compared to state-of-the-art approaches. This demonstrates the effectiveness and efficiency of our proposed method.

\subsection{Ablation Study}
In the ablation study, we use a shorter training schedule.  

\begin{table*}
	\caption{Ablation study. Default settings are marked in \colorbox{gray!20}{gray}.}
	\begin{subtable}[t]{0.495\linewidth}
		\centering
		\setlength{\tabcolsep}{5mm}
		{
			\begin{tabular}{ccc|cc}
				\toprule
				TDA & MISE & ALT &mAP  &NDS \\
				\midrule
				-  &- &- &67.17 &70.39\\
				$\surd$  &- &- &67.90 &70.54\\
				$\surd$  &$\surd$ &- &68.55 &71.18\\
				\rowcolor{gray!20}
				$\surd$  &$\surd$ &$\surd$ &69.14 &71.91\\
				\bottomrule
		\end{tabular}}%
            \vspace{5pt}
		\caption{Module ablation}
	\end{subtable}
	\begin{subtable}[t]{0.495\linewidth}
		\centering
		\setlength{\tabcolsep}{4mm}{
			\begin{tabular}{cc|cc}
				\toprule
				$\Psi(C) \backsim \Psi(L)$ & $\Psi(C,L) \backsim \Psi(G)$& mAP &NDS \\
				\midrule
				- &- &67.86 &71.05 \\
				$\surd$  &- &68.18 &71.19 \\
				-  &$\surd$ &68.58 &71.52 \\
				\rowcolor{gray!20}
				$\surd$  &$\surd$ &69.14 &71.91 \\
				\bottomrule
		\end{tabular}}
              \vspace{5pt}
		\caption{Domain aligning}
	\end{subtable}
	\begin{subtable}[t]{0.495\linewidth}
		\centering
		\setlength{\tabcolsep}{7.6mm}{
			\begin{tabular}{cc|cc}
				\toprule
				MI & SE& mAP &NDS \\
				\midrule
				- &- &68.48 &70.89  \\
				$\surd$  &- &68.67 &71.62 \\
				-  &$\surd$ &68.63 &71.57 \\
				\rowcolor{gray!20}
				$\surd$  &$\surd$ &69.14 &71.91 \\
				\bottomrule
		\end{tabular}}
              \vspace{5pt}
		\caption{Modal interaction and specialty enhancement}

	\end{subtable}
	\begin{subtable}[t]{0.495\linewidth}
		\centering
		\setlength{\tabcolsep}{7.8mm}
		{
			\begin{tabular}{cc|cc}
				\toprule
				Cls & IoU& mAP &NDS \\
				\midrule
				- &- &68.55  &71.18   \\
				$\surd$  &- &68.77 &71.58 \\
				-  &$\surd$ &68.73 &71.42 \\
				\rowcolor{gray!20}
				$\surd$  &$\surd$ &69.14 &71.91 \\
				\bottomrule
		\end{tabular}}%
              \vspace{5pt}
		\caption{Mode of ATL}
	\end{subtable}
	\begin{subtable}[t]{0.495\linewidth}
		\centering
		\setlength{\tabcolsep}{9mm}
		{
			\begin{tabular}{c|cc}
				\toprule
				Interaction & mAP &NDS \\
				\midrule
				Global  &67.81 &70.78 \\
				Local   &68.04 &70.53 \\
				\rowcolor{gray!20}
				Deformable  &69.14 &71.91 \\
				\bottomrule
		\end{tabular}}%
                \vspace{5pt}
		\caption{Modal interaction encoding manner.}
	\end{subtable}
	\begin{subtable}[t]{0.495\linewidth}
		\centering
		\setlength{\tabcolsep}{8.3mm}
		{
			\begin{tabular}{c|cc}
				\toprule
				Encoder numbers & mAP &NDS \\
				\midrule
				%			1   &68.66 &71.70 \\
				2   &68.89 &71.86 \\
				\rowcolor{gray!20}
				3   &69.14 &71.91 \\
				4	&68.90 &71.71 \\
				%			5   &68.79 &71.59 \\
				\bottomrule
		\end{tabular}}%
            \vspace{5pt}
		\caption{Numbers of TDA encoder}
	\end{subtable}
	\begin{subtable}[t]{0.495\linewidth}
		\centering
		\setlength{\tabcolsep}{8.6mm}
		{
			\begin{tabular}{c|cc}
				\toprule
				Weight  & mAP &NDS \\
				\midrule
				Share weight  &68.64 &71.81 \\
				\rowcolor{gray!20}
				Specific  weight  &69.14 &71.91 \\
				\bottomrule
		\end{tabular}}%
                \vspace{5pt}
		\caption{Specialty enhancement weight manner}
	\end{subtable}
	\begin{subtable}[t]{0.506\linewidth}
		\centering
		\setlength{\tabcolsep}{8.1mm}
		{
			\begin{tabular}{c|cc}
				\toprule
				Perception order & mAP &NDS \\
				\midrule
				Space $\rightarrow$ Channel   &68.85 &71.90 \\
				\rowcolor{gray!20}
				Channel $\rightarrow$ Space   &69.14 &71.91 \\
				\bottomrule
		\end{tabular}}%
                \vspace{5pt}
		\caption{Dynamic fusion order}
	\end{subtable}
	\label{abla}
       \vspace{-10pt}
\end{table*}

\subsubsection{The Effectiveness of Framework}
As shown in Table \ref{abla}-a, the first row represents our baseline. We then introduce the Triphase Domain Aligning (TDA) module, which adjusts the modal distribution and aligns them. With this enhancement, we achieve 67.9\% mAP and 70.54\% NDS. 

Next, we integrate the Modal Interaction and Specialty Enhancement (MISE) module. This module investigates the correlation between two modalities and enhances the object region features in terms of perception. With MISE, we achieve a mAP of 68.55\% and a NDS of 71.18\%.

Finally, we introduce the Adaptive Learning Technique (ALT), which fuses semantics and geometry information to enhance instance training. ALT yields a mAP of 69.14\% and a NDS of 71.91\%. The results demonstrate that each module contributes to improving the 3D detection flow.

\subsubsection{The Effectiveness of TDA} 
In Table \ref{abla}-b, we observe that $\Psi(C)\backsim\Psi(L)$ represents the alignment between the camera and LiDAR feature domains. Additionally, $\Psi(C, L)\backsim\Psi(G)$ denotes the alignment of camera and LiDAR features with the ground truth domain. Comparing single alignment and non-alignment approaches, the triphase domain alignment achieves the best performance, with a mAP of 69.14\% and a NDS of 71.91\%. This alignment method effectively brings the two modalities closer to each other in space distribution, aligning them with the gt.

For efficiency considerations, we employ a CNN as the encoder to obtain the aligned multi-modal feature. In Table \ref{abla}-f, we incrementally vary the number of encoders from two to four. When the aligning encoder has too few layers, it fails to effectively adjust feature distributions. Conversely, if the aligning encoder has too many layers, it risks overfitting the alignment operation, leading to suboptimal performance. Consequently, we settle on a three-layer structure finally.

\subsubsection{The Effectiveness of MISE} 
In this section, we delve into the design of each component within the multi-modal fusion module. During modal interaction representation learning (MI), we employ a transformer encoder to capture the relevance between the camera and LiDAR modalities, enhancing their similar representations. As demonstrated in Table \ref{abla}-e, we explore three interaction modes:

1. The $Global$ mode computes interactions across all points, resulting in a mean average precision of 67.81\%.

2. The $Local$ mode restricts interactions to only nine points in the vicinity, achieving a mAP of 68.04\%.

3. The $Deformable$ mode considers nearby points while allowing learnable adjacency point offsets. This mode achieves the highest performance, with a mAP of 69.14\%.

The results underscore the superiority of the $Deformable$.

Next, we delve into the specialty enhancement module. We model each feature point using an error distribution diagram. When a point exhibits a low offset from the ground truth center and low uncertainty, it yields a high perception degree. This enhanced perception contributes to improving the representation of each modality's features. In Table \ref{abla}-g, we investigate whether to share encoding weights between the two modalities. The results demonstrate that using specific weights for each modality leads to better performance.

As shown in Table \ref{abla}-c, we evaluate the whole process of modal interaction and specialty enhancement. In the first and second rows, each of them can enhance model performance (68.67\%/ 68.63\%mAP). And the combination of them (69.14\% mAP) can sufficiently dig the potential information and dynamically fuse features.

Finally, in the dynamic fusion process, we explore two different perception orders, as outlined in Table \ref{abla}-h. Our default operation involves dynamically fusing the channels first and then fusing them in space, resulting in a mAP of 69.14\%. Additionally, we test an alternative perception order: space perception first, followed by channel perception (yielding a mAP of 68.85\%). The results clearly favor the default setting.

\subsubsection{The Effectiveness of ALT} 
Our adaptive learning technique departs from the traditional equal treatment of different objects by dynamically assigning instance-specific adaptive weights during optimization. In our default setting, we incorporate both classification confidence and the IoU joint score as perception loss weights. As demonstrated in Table \ref{abla}-d, we also explore alternative strategies, such as using only the classification confidence or IoU as the index. The results indicate that the classification-only strategy achieves a mean average precision (mAP) of 68.77\%, while the IoU-only mechanism yields a slightly lower mAP of 68.73\%.

% new figure
\begin{figure*}
	\centering
	\includegraphics[scale=0.14]{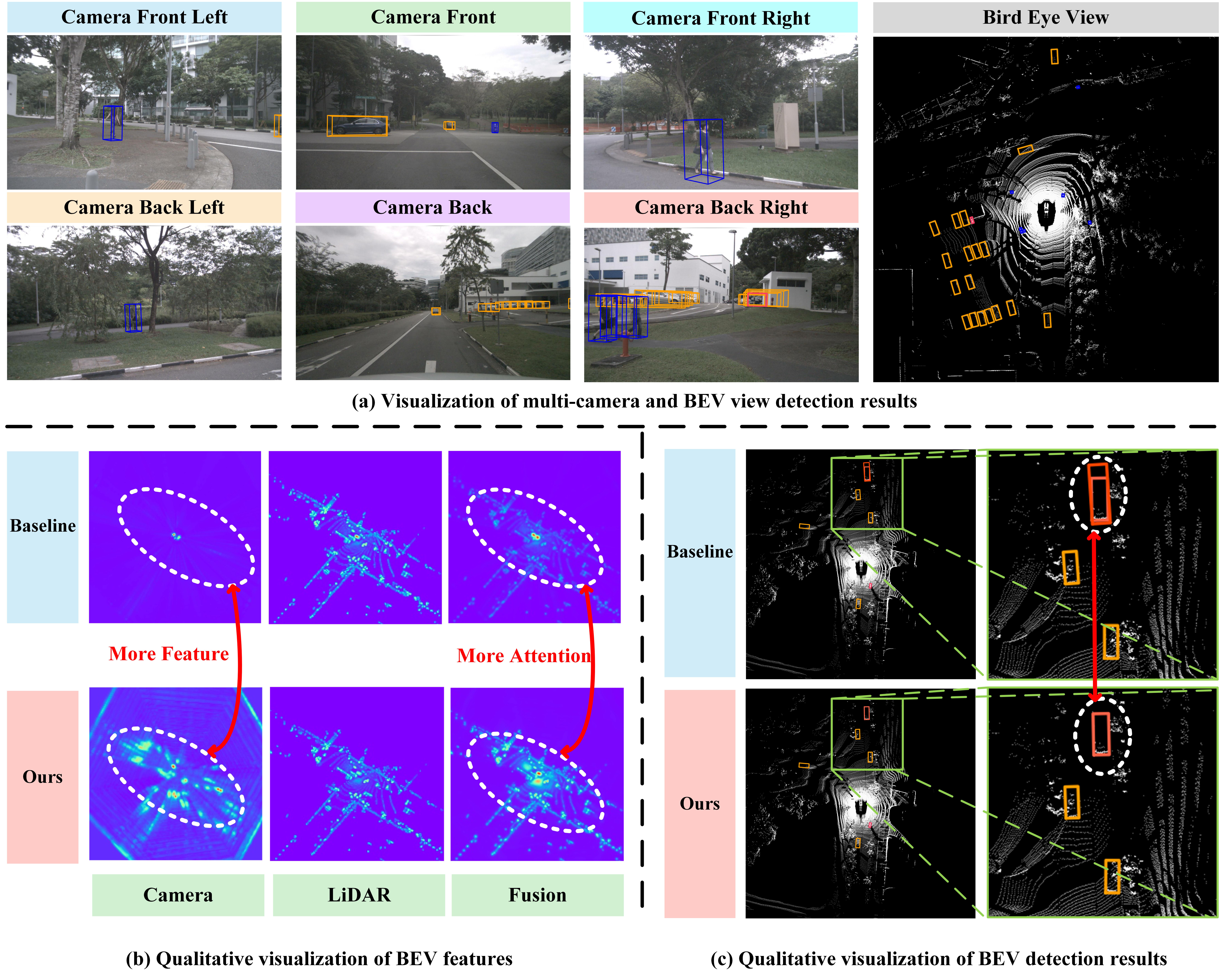}
	\caption{Visualization.}
	\label{vis}
        \vspace{-10pt}
\end{figure*}

% original figure 
% \begin{figure*}
% 	\centering
% 	\includegraphics[scale=0.095]{./fig/vis0.jpg}
% 	\caption{Visualization of multi-camera and BEV view detection results. }
% 	\label{vis_det}
% \end{figure*}

% \begin{figure}
% 	\centering
% 	\includegraphics[scale=0.072]{./fig/viz5.jpg}
% 	\caption{Qualitative visualization of BEV features. Our model retain the representation of camera modality and enhance the fusion representation ability.}
% 	\label{vis_feature}
% \end{figure}

% \begin{figure}[h]
% 	\centering
% 	\includegraphics[scale=0.08]{./fig/viz4.jpg}
% 	\caption{Qualitative visualization of BEV detection results. Our model can better avoid the overlap results.}
% 	\label{vis_bev}
% \end{figure}

\subsubsection{Visualization}
In this section, we analyze the model performance from a qualitative perspective. 

First, as depicted in Fig. \ref{vis} (a), we present visualizations from both the multi-camera and bird’s-eye-view (BEV) perspectives. The left part demonstrates accurate detection across most categories from each direction. Meanwhile, the right part showcases BEV detection results, providing an intuitive representation of peripheral objects.

% Furthermore, as depicted in Fig. \ref{vis_feature}, our model not only preserves the representation from the camera modality but also enhances fusion representation capabilities compared to the baseline, which predominantly relies on LiDAR modality.

Moreover, as depicted in Fig. \ref{vis} (b), we compare the features of multi-camera, LiDAR, and fusion modes between the baseline and our model (both of which are well-trained). In the first row, the baseline model heavily relies on the LiDAR modality. While its fusion features closely resemble those of LiDAR, the camera features are nearly absent, indicating that some information is neglected. The results indicate that the model tends to optimize for easy learning while neglecting the more challenging aspects. This behavior may not be suitable for complex environments. In contrast, our model, shown in the second row, maintains diverse features from both the camera and LiDAR modalities. This complementary combination allows for richer fusion features. In summary, our model not only preserves representations from the camera modality but also enhances fusion capabilities compared to the baseline, which predominantly relies on LiDAR.

Furthermore, as depicted in Fig. \ref{vis} (c), we present visualizations comparing the bird’s-eye-view (BEV) results between the baseline and our method. Different color bounding boxes represent diverse object categories. We observe that the baseline often detects multiple overlapping objects (with high Intersection over Union, or IoU, between them) at the same location, which is unrealistic in the real world (two cars would not overlap on the road). In contrast, our method dynamically fuses better features and trained instances, leading to improved handling of such situations.

\section{Conclusion}
In this paper, we explore the LiDAR-camera dynamic adjustment fusion framework for 3D object detection. First, a triphase domain aligning module is introduced to help adjust the distribution of multimodal features. Moreover, we propose to interact with multi-modal and enhance their specialty. A dynamic fusion strategy is designed to fuse the features with space and channel perspective. Finally, we adopt an adaptive learning technique that achieves semantics and geometry information aggregation and dynamically optimizes diverse instances. To validate the effectiveness of our proposed framework, we conducted extensive experiments using the nuScenes dataset. The results demonstrate significant improvements over existing methods, highlighting the practical utility of our approach.

\bibliographystyle{splncs04}
\bibliography{main}

\addtolength{\textheight}{-12cm}   % This command serves to balance the column lengths

%%%%%%%%%%%%%%%%%%%%%%%%%%%%%%%%%%%%%%%%%%%%%%%%%%%%%%%%%%%%%%%%%%%%%%%%%%%%%%%%

%%%%%%%%%%%%%%%%%%%%%%%%%%%%%%%%%%%%%%%%%%%%%%%%%%%%%%%%%%%%%%%%%%%%%%%%%%%%%%%%



% \section*{ACKNOWLEDGMENT}

% The preferred spelling of the word ÒacknowledgmentÓ in America is without an ÒeÓ after the ÒgÓ. Avoid the stilted expression, ÒOne of us (R. B. G.) thanks . . .Ó  Instead, try ÒR. B. G. thanksÓ. Put sponsor acknowledgments in the unnumbered footnote on the first page.

%%%%%%%%%%%%%%%%%%%%%%%%%%%%%%%%%%%%%%%%%%%%%%%%%%%%%%%%%%%%%%%%%%%%%%%%%%%%%%%%

% References are important to the reader; therefore, each citation must be complete and correct. If at all possible, references should be commonly available publications.

\end{document}